%% file: main.tex
\documentclass[letterpaper, 10 pt, conference]{ieeeconf}  

\IEEEoverridecommandlockouts                              

\usepackage{comment}
\usepackage{multicol}
\usepackage{multirow}
\usepackage{mathtools}
\usepackage{booktabs}
\usepackage{graphicx} 
\usepackage{xcolor,soul}
\usepackage{times} 
\usepackage{amsmath} 
\usepackage{amssymb}  
\usepackage{caption}
\usepackage{pifont}
\usepackage{hyperref}
\usepackage{subcaption}
\usepackage{float}
\usepackage[font=footnotesize]{caption} 
\usepackage{soul}
\makeatletter
\g@addto@macro\normalsize{%
  \setlength\abovedisplayskip{4pt}
  \setlength\belowdisplayskip{4pt}
  \setlength\abovedisplayshortskip{4pt}
  \setlength\belowdisplayshortskip{4pt}
}
\makeatother

\title{
    \LARGE \bf
    Map-Aware Human Pose Prediction for Robot Follow-Ahead
}

\author{
    Qingyuan Jiang, Burak Susam, Jun-Jee Chao and Volkan Isler\\
    University of Minnesota\\
    Shepherd Laboratories, 100 Union St SE\\
    {\tt\small \{jian0345, susam001, chao0107, isler\}@umn.edu} 
}

\begin{document}

\twocolumn[{
\renewcommand\twocolumn[1][]{#1}%
\maketitle
\begin{center}
        \centering
            \includegraphics[width=1.0\textwidth]{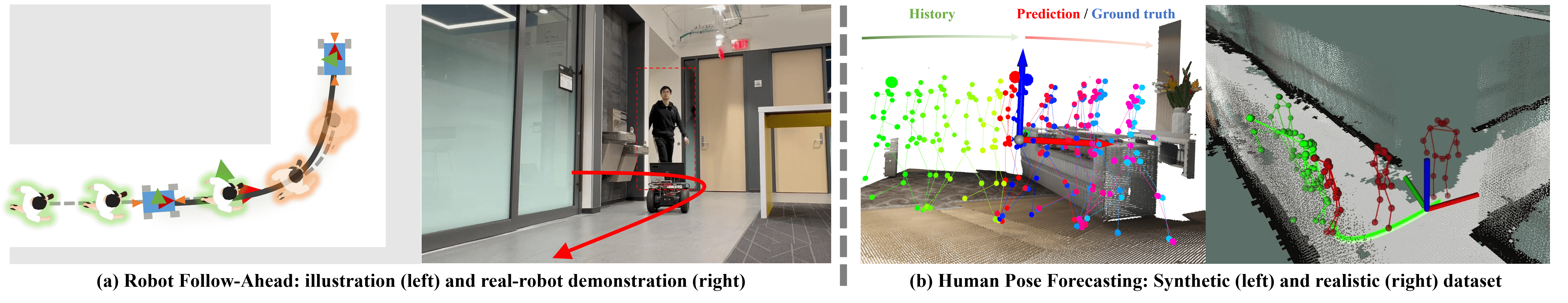}
            \captionof{figure}
            {
                \textbf{(a) The Robot Follow-Ahead Task:} 
                A mobile robot maintains the sight of a human actor while driving in front of them in an indoor environment 
                \textbf{(b) Map-aware human pose prediction.} 
                To achieve the follow-ahead task, given the pose histories (shown in green), we predict long-term human poses (shown in red with ground truth in blue) by incorporating the local map information and generating input for a predictive robot controller.
            }
\end{center}
}]

\thispagestyle{empty}
\pagestyle{empty}


\input{sections/00-abstract}

\section{INTRODUCTION}  \label{sec: introduction}
\input{sections/01-introduction}

\section{RELATED WORK} \label{sec: related_work}
\input{sections/02-related_work}

\section{PROBLEM FORMULATION}  \label{sec: formulation}
\input{sections/03-formulation}

\section{APPROACH} \label{sec: approach}

\input{sections/04-approach}

\section{SYSTEM DESIGN} \label{sec: system}

\input{sections/05-system}

\section{DATASET} \label{sec: dataset}
\input{sections/06-dataset}

\section{EXPERIMENTS} \label{sec: experiments}
\input{sections/07-experiments}

\section{CONCLUSION}    \label{sec: conclusion}
\input{sections/08-conclusions}



\bibliographystyle{unsrt}
\bibliography{main}

\end{document}


\maketitle

\input{sections/supplementary/intro}

\section{Robot Follow-Ahead}    \label{appendix: follow-ahead}
\input{sections/supplementary/follow-ahead}

\section{Robot System}    \label{appendix: robot}
\input{sections/supplementary/robot}

\section{Dataset}   \label{appendix: dataset}
\input{sections/supplementary/dataset}


\bibliographystyle{unsrt}
\bibliography{ref}

%% file: sections/00-abstract.tex
\begin{abstract}
In the robot follow-ahead task, a mobile robot is tasked to maintain its relative position in front of a moving human actor while keeping the actor in sight. To accomplish this task, it is important that the robot understand the full 3D pose of the human (since the head orientation can be different than the torso) and predict future human poses so as to plan accordingly. This prediction task is especially tricky in a complex environment with junctions and multiple corridors. In this work, we address the problem of forecasting the full 3D trajectory of a human in such environments. Our main insight is to show that one can first predict the 2D trajectory and then estimate the full 3D trajectory by conditioning the estimator on the predicted 2D trajectory. With this approach, we achieve results comparable or better than the state-of-the-art methods three times faster. As part of our contribution, we present a new dataset where, in contrast to existing datasets, the human motion is in a much larger area than a single room. We also present a complete robot system that integrates our human pose forecasting network on the mobile robot to enable real-time robot follow-ahead and present results from real-world experiments in multiple buildings on campus. Our project page, including supplementary material and videos, can be found at: \url{https://qingyuan-jiang.github.io/iros2024\_poseForecasting/}

\end{abstract}

%% file: sections/01-introduction.tex
Imagine a robot working as a photographer and recording videos in front of a moving actor. To keep sight of the front of the actor's body and facial expressions, the robot needs to drive in front of the actor, catch their pace, and actively predict their motion while avoiding obstacles in the environment. 
This task of controlling a robot to maintain the visibility of a human actor while driving in front of them, is called ``robot follow-ahead"~\cite{ho_behavior_2012, karnad_modeling_2012}.
A key component of existing robot follow-ahead approaches is to predict future human poses~\cite{mahdavian_stpotr_2022} to help the robot stay in front of the actor without losing their sight.


While arbitrary human trajectories are feasible in open spaces, obstacles in indoor environments constrain the set of possible motions and, therefore, reduce the space of available trajectories.
To provide sufficient information for human pose forecasting, researchers have been exploring the advantage of using environmental information~\cite{hassan_stochastic_2021, wang_geometric_2021, wang_scene-aware_2021, huang_diffusion-based_2023}. 
Suppose a human is walking in a hallway facing a left turn ahead. With high probability, the actor would take a left turn after a few steps. 
Such information may help with human pose prediction over a long time horizon. 
Aiming for the follow-ahead task in a real-robot setting, we show how environment information can be used to predict long-horizon human poses and propose a real-time human pose prediction method that runs three times faster than the state-of-the-art methods and performs better or comparable.

\begin{figure}[t]
    \centering
    \includegraphics[width=0.95\columnwidth]{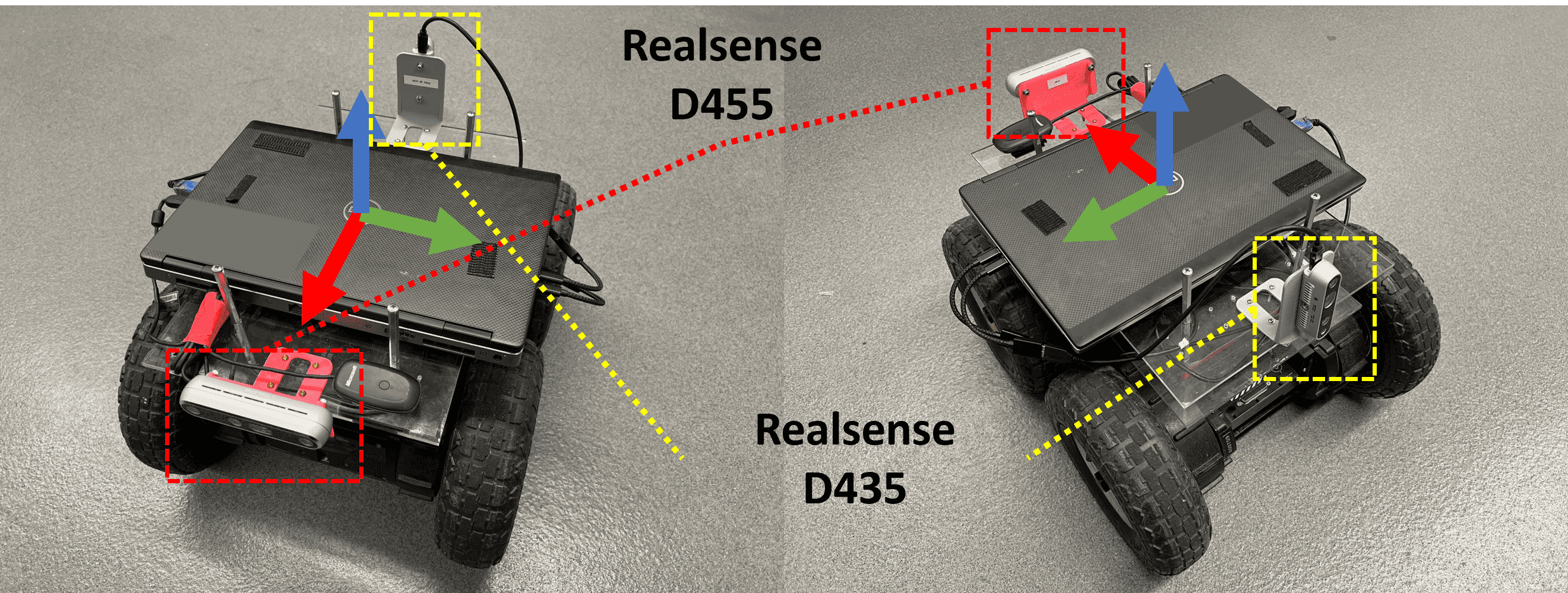}
    \caption{\textbf{Robot}. Our mobile robot is assembled with two Realsense RGB-D cameras based on a Rover robot. We use the front camera to build the map for localization while navigating. The rear camera detects and tracks the human actor for 3D skeleton poses. The robot coordinate frame is also shown.
    }
    \label{fig: hardware}
    \vspace{-6pt}
\end{figure}

Moreover, existing human pose forecasting datasets that contain environmental information are either limited to a single room region such as PROX~\cite{hassan_resolving_2019} or gathered from synthetic data~\cite{cao_long-term_2020}. To overcome these issues, we built a robot (Fig.~\ref{fig: hardware}) with two cameras that can localize the robot and simultaneously record human motion in a building-scale space. We collect a realistic dataset (Real-IM: Real Indoor Motion Dataset) containing human motion with complete environmental information and multiple human motion styles.
We train our human pose prediction model on both synthetic data~\cite{cao_long-term_2020} and on our new Real-IM dataset. 
We show in Sec.~\ref{sec: experiments} that our method outperforms the state-of-the-art methods for predicting both the trajectory and pose of the human. 
With the predicted human pose, we demonstrate that the proposed real-time algorithm enables the robot to follow a human in the front in real-world experiments. 

From the robot follow-ahead perspective, the closest work is~\cite{mahdavian_stpotr_2022}, which uses human pose prediction results to perform robot follow-ahead in an open space. 
However, they require an extra third-person-view camera to localize the actor and are therefore constrained in a single-room region. 
Ours is the first work to demonstrate robot follow-ahead in a building-scale environment without relying on other off-board cameras and utilizing full 3D pose information. 

Our contributions can be summarized as follows:

\begin{enumerate}
    \item We present a new real-time method for human pose forecasting that considers the surrounding environment represented as occupancy maps. 
    \item We present a new building-scale real-world dataset for human pose prediction by building a robot system with dual cameras for localization and navigation.
    \item We conduct experiments on synthetic and the proposed realistic datasets. Our method outperforms the baselines and state-of-the-art methods on both trajectory prediction and long-term human pose forecasting. Moreover, we show that our method is three times faster than existing methods, which allows us to perform robot follow-ahead in a real-world setup with a simple controller. 
\end{enumerate}

%% file: sections/02-related_work.tex
We summarize related work in two directions: robot follow-ahead and the human pose prediction.

\subsection{Robot Follow-Ahead}
The human follow-ahead problem was studied earlier in~\cite{ho_behavior_2012}. 
Kalman Filter~\cite{ho_behavior_2012} or Extended Kalman Filter~\cite{nikdel_hands-free_2018, oh_chance-constrained_2015} are used to estimate the human trajectory. 
The Deep Reinforcement Learning method in~\cite{nikdel_lbgp_2021} implicitly predicts the human trajectory by setting rewards. None of these works predicted full human poses. 
Recently, the work of~\cite{mahdavian_stpotr_2022} jointly predicts the human trajectory and human poses for the robot to follow ahead in an open space. However, an offboard third-person-view camera is still needed. 
In addition, human tractories are predicted in~\cite{mangalam_goals_2021, djuric_uncertainty-aware_2020} in Bird-Eye view (BEV). One recent work~\cite{salzmann_robots_2023} predicts the human trajectory using the human skeleton poses. Our work focuses on human pose prediction for the robot's follow-ahead problem and proposes to plan the robot's path by forecasting the human pose given the environmental information.

\subsection{Human Pose Prediction} 

Prediction of the human pose attracted increasing attention in the past few years~\cite{lyu_3d_2022, saadatnejad_generic_2023, nikdel_dmmgan_2023, yuan_dlow_2020}. 
Researchers have managed to predict the human poses in an open space using architectures such as Multi-Layer Perception (MLP)~\cite{bouazizi_motionmixer_2022}, Generative Adversarial Networks (GANs)~\cite{chopin_3d_2022, nikdel_dmmgan_2023}, diffusion-based method~\cite{saadatnejad_generic_2023}, Graph Convolutional Networks (GCN)~\cite{mao_learning_2019},  or Transformers~\cite{chen_sttg-net_2022, lucas_posegpt_2022, martinez-gonzalez_pose_2021, aksan_spatio-temporal_2021}.

Recent work started to incorporate environmental information with the human motion prediction task~\cite{cao_long-term_2020, mao_contact-aware_2022, hassan_stochastic_2021} and the human pose synthesis task~\cite{wang_towards_2022, wang_scene-aware_2021, hassan_resolving_2019, wang_synthesizing_2021}. 
This body of literature relies on data generated with a simulation-based large-scale dataset~\cite{cao_long-term_2020} and in PROX~\cite{hassan_resolving_2019} with a single-room realistic dataset. 
In order to deal with the human-environment interaction, some work explicitly predicts the goal and the path (or contact points)~\cite{cao_long-term_2020, hassan_stochastic_2021, mao_contact-aware_2022}. Methods include using conditional variational autoencoders (cVAE)~\cite{cao_long-term_2020, hassan_stochastic_2021} or GANs~\cite{wang_scene-aware_2021}.
Computationally expensive calculations such as voxelization~\cite{hassan_stochastic_2021} or contact map~\cite{mao_contact-aware_2022} are involved.
Some work extract the environment features implicitly using PointNet~\cite{wang_towards_2022, wang_synthesizing_2021} or ResNet18~\cite{wang_scene-aware_2021, wang_geometric_2021}. 
Meanwhile, diffusion-based models have been proposed to generate human poses~\cite{huang_diffusion-based_2023}.

We build our work on~\cite{mao_contact-aware_2022}. We use a similar GRU-based network structure, but we focus on different scales of the human pose forecasting problem. 
By considering the interaction between the full 3D environment point cloud and the actor's joints, CA~\cite{mao_contact-aware_2022} addresses better on the complex human-environment interaction in a small space, such as sitting on a sofa or lying in bed. 
In contrast, we aim to forecast human poses for robot follow-ahead tasks. 
We focus on larger indoor scenes while the human is walking with less contact with objects. 
We improve and speed up the long-term trajectory prediction by conditioning it on a 2D occupancy map.
In Sec.~\ref{sec: exp_pose}, we show that considering only 2D maps is faster than considering the full 3D geometry for this setup and can achieve a better forecasting performance in a long time horizon.

%% file: sections/03-formulation.tex
We formulate the human pose prediction problem in a known environment as follows. 
We define a human pose with $J$ joints as $\mathbf{x} \in \mathbb{R}^{J \times 3}$. The past $N$-step human motion is represented as $\mathbf{X}_{1:N} = [\mathbf{x}_1, \mathbf{x}_2, \ldots, \mathbf{x}_N] \in \mathbb{R}^{N \times J \times 3}$. 
All the human joints are represented in the latest human pose's coordinate frame.
We use the human actor's torso (hip) keypoint's position $\mathbf{x}^a$ to represent the human trajectory in 2D.
We denote the human's surrounding environment as $\mathbf{S}$. 
Given the human pose history $\mathbf{X}_{1:N}$ and the surrounding environment $\mathbf{S}$, we would like to forecast the future human poses $\mathbf{X}_{N+1:N+T}$ with our network $\mathcal{F}$ within time horizon $T$. That is:

\begin{equation}
    \mathbf{\hat{X}}_{N+1:N+T} = \mathcal{F}(\mathbf{X}_{1:N}, \mathbf{S})
\end{equation}



Similar to existing works, we assume only one person is observed by the robot. 
We also assume the environment is entirely static and known to the robot. 
In Sec.~\ref{sec: experiments}, we investigate the performance of human pose prediction given a fully known map, a limited field of view, and a fully unknown map.

%% file: sections/04-approach.tex
\begin{figure*}[th]
    \centering
    \includegraphics[trim={1cm 3cm 0cm 5.2cm},clip,width=\textwidth]{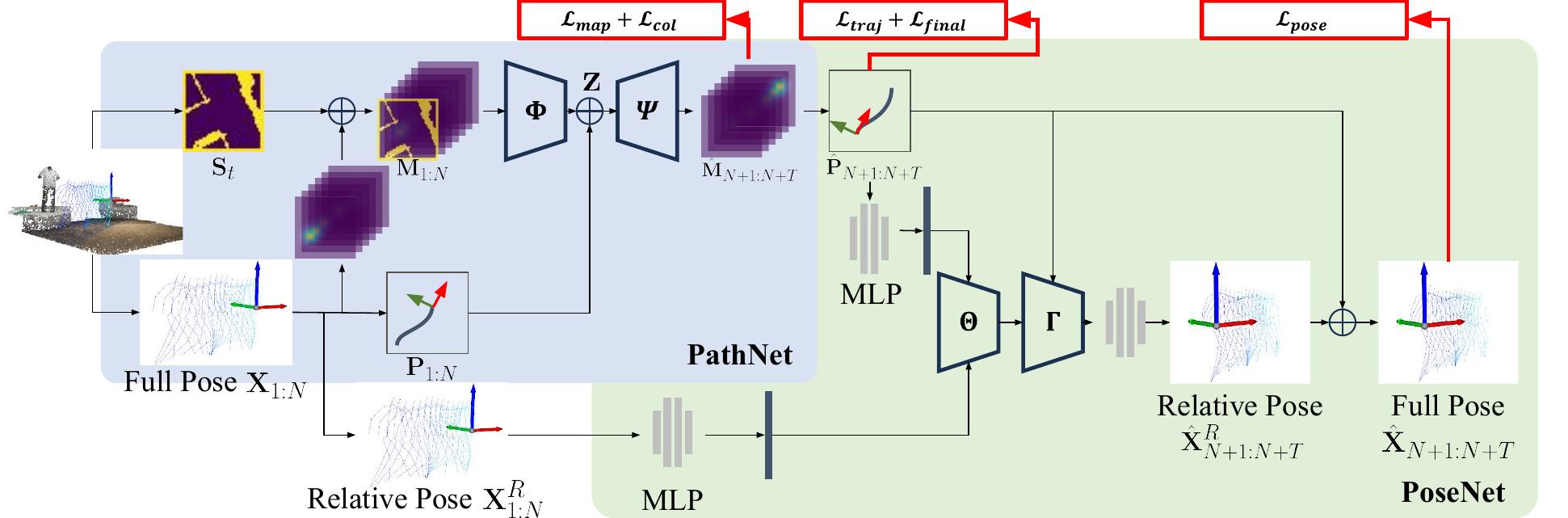}
    \caption{\textbf{Network}. Our network has two parts. A PathNet to predict human trajectory, and a PoseNet to predict human future poses. The PathNet takes input from the occupancy map as well as the human trajectory and predicts the human future trajectory. The PoseNet uses the prediction results and local pose as input and predicts the future poses with a Gated Recurrent Unit (GRU) based network.}
    \label{fig: network}
    \vspace{-5pt}
\end{figure*}

Given human history poses and surrounding environment, our method predicts the future poses in two steps similar to~\cite{cao_long-term_2020, hassan_stochastic_2021}: first predict a human trajectory in the 2D environment, then complete the full-body pose based on the predicted trajectory. 
Note that the human heading direction can differ from the torso and the 2D trajectory. 
The motion planning module needs a full-body pose as input to calculate the viewing quality. 
Therefore, we predict full-body poses instead of a 2D trajectory for human motion.
The following subsections introduce our representation of the environment and human poses. Then, we describe each component in our network architecture in detail.

\subsection{Representation} \label{sec: representation}

\textbf{Occupancy Map}. We want to use environmental information to help with human motion prediction. We represent the environment with the local occupancy map $\mathbf{S}_t$ around the human pose $\mathbf{X}_t$. We clip the local occupancy map by distance $d_x$ and $d_y$ along $x$-axis and $y$-axis of the actor frame, respectively. We denote the resolution of the occupancy map as $r$.

We encode the human pose $\textbf{X}_{1:N}$ with two more representations. A trajectory map and a local pose.

\textbf{Trajectory and trajectory map}.  First, we extract the trajectory of the human based on the torso, i.e., $\mathbf{x}_1^a, \mathbf{x}_2^a, \ldots, \mathbf{x}_N^a$. We denote the 2D human trajectory path as $\mathbf{P}_{1:N} = (\mathbf{x}_1^a, \mathbf{x}_2^a, \ldots, \mathbf{x}_N^a) \in \mathbb{R}^{N \times 2}$.
Then, we encode them into a trajectory map. At each timestamp $\mathbf{x}_t^a = (x, y, z)$, we project $\mathbf{x}_t^a$ into the 2D map with the same resolution and center as the local occupancy map. We use a binary map as the ground truth for training. $I(u,v) = 1$ if $(u, v) = ((x, y) - (d_x, d_y)) / r$. $u$, $v$ is the pixel coordinate. We use a Gaussian distribution heatmap as the observation. The Gaussian has a center at the projected position with a predefined covariance $\sigma$. $I(u,v) = \mathcal{N}((u, v), \sigma)$. Thus, we represent the human torso trajectory with the trajectory map $\mathbf{M_{1:N}}$ by shape $H \times W \times N$, as shown in Fig.~\ref{fig: network}.

\textbf{Local pose}. For a human pose $\mathbf{x}$, we subtract the human pose by the torso position $\mathbf{x}^a$ to describe the human pose regardless of the position. We denote it as $\mathbf{x}^R = \mathbf{x} - \mathbf{x}^a$. And $\mathbf{X}^R_{1:N} = [\mathbf{x}^R_{1}, \mathbf{x}^R_{2}, \ldots, \mathbf{x}^R_{N}]$.

\subsection{PathNet}    \label{sec: pathnet}

We predict the human trajectory with the occupancy map and the human trajectory history based on a U-net~\cite{ronneberger_u-net_2015} architecture similar to~\cite{mangalam_goals_2021}. We concatenate the occupancy map of the last frame $\mathbf{S}_N$ and the human past trajectory map $\mathbf{M}_{1:N}$, and input it to an encoder $\mathbf{\Phi}$. We extract the bottleneck vector as the latent feature vector $\mathbf{z}$ and concatenate it with the past trajectory $\mathbf{P}_{1:N}$. We decode the feature vector with the decoder $\mathcal{\psi}$ to a probability trajectory map $\hat{\mathbf{M}}_{N+1:N+T}$ with shape $H \times W \times T$. We use soft-argmax~\cite{sun_integral_2018, luvizon_human_2017} to calculate the actor's position for each time stamp in the future.

\begin{align}
    \mathbf{z}_{traj} &= \mathbf{\Phi}(\mathbf{S}_N, \mathbf{M}_{1:N})  \\
    \hat{\mathbf{M}}_{N+1:N+T} &= \mathbf{\Psi}(\mathbf{z}_{traj}, \mathbf{P}_{1:N})  \\
    \hat{\mathbf{P}}_{N+1:N+T} &= \text{soft}\arg\max (\hat{\mathbf{M}}_{N+1:N+T})
\end{align}

We use multiple loss functions as training objectives. We define a trajectory loss $\mathcal{L}_{traj}$, a final position loss $\mathcal{L}_{final}$, and a trajectory map loss $\mathcal{L}_{map}$ and a collision loss $\mathcal{L}_{col}$.

\begin{equation}
    \begin{split}
        \mathcal{L}_{traj} &= \frac{1}{T} \sum_{t=N+1}^{N+T} \|\mathbf{P}_t - \hat{\mathbf{P}}_t \|_2 \\
        \mathcal{L}_{final} &= \|\mathbf{P}_{N+T} - \hat{\mathbf{P}}_{N+T}\|_2  \\
        \mathcal{L}_{map} &= \frac{1}{T} \sum_{t=N+1}^{N+T} \text{BCE}(\mathbf{M}_{t}, \hat{\mathbf{M}}_t) \\
        \mathcal{L}_{col} &= \frac{1}{T} \sum_{t=N+1}^{N+T} (\hat{\mathbf{M}}_t \cdot |\mathbf{S}|)
    \end{split}
\end{equation}

$\text{BCE}$ represents the Binary Cross-Entropy loss with weight $w$. The overall loss term is given by the sum of the loss terms with weights $\lambda$. In the experiment section, we empirically select $w=40$, $\lambda_3 = 2$ and $\lambda_1 = \lambda_2 = \lambda_4 = 1.0$.

\begin{equation}
    \mathcal{L} = \lambda_1 \mathcal{L}_{traj} + \lambda_2 \mathcal{L}_{final} + \lambda_3 \mathcal{L}_{map} + \lambda_4 \mathcal{L}_{col}
\end{equation}

\subsection{PoseNet}    \label{sec: posenet}

Given the trajectory prediction $\hat{\mathbf{P}}_{N+1:N+T}$ and human local pose history $\mathbf{X}_{1:N}^R$, we predict the human local pose $\hat{\mathbf{X}}_{N+1:N+T}^R$ with a Gated Recurrent Unit (GRU) network first. Then, we transform the local pose to the predicted position by each time stamp. We first encode the past local pose $\mathbf{X}_{1:N}^R$ and predicted trajectory $\hat{\mathbf{P}}_{N+1:N+T}$ separately with Multi-layer Perceptrons (MLPs). We concatenate and feed the latent feature $\mathbf{z}_{pose}$ into the GRU network $\mathbf{\Theta}$. We then decode the local pose recurrently by a GRU cell module $\mathbf{\Gamma}$ using the latent feature $\mathbf{z}_{pose}$ at each time stamp.

\begin{equation}
    \begin{split}
        &\mathbf{z}_{pose, N} = \mathbf{\Theta}(\mathbf{X}_{1:N}^R, \hat{\mathbf{P}}_{N+1:N+T}) \\
        &\mathbf{z}_{pose, (t+1)}, \quad \hat{\mathbf{X}}_{t+1}^R = \mathbf{\Gamma}(\mathbf{z}_{pose, t}, \hat{\mathbf{P}}_t)   \\
        &\hat{\mathbf{X}}_{t} = \hat{\mathbf{X}}_{t}^R + \hat{\mathbf{P}}_t        
    \end{split}
\end{equation}

The loss function $\mathcal{L}_{pose}$ is given in Eq.~\ref{eq: loss_pose}. We use the ground truth trajectory as the input during the training time and trajectory prediction from the PoseNet at inference time.

\begin{equation}    \label{eq: loss_pose}
    \mathcal{L}_{pose} = \frac{1}{TJ} \sum_{t=N+1}^{N+T} \sum_{J} \|\mathbf{X}_t^j - \hat{\mathbf{X}}_t^j \|_2
\end{equation}



%% file: sections/05-system.tex
We first present the mobile robot system in Sec.~\ref{sec: hardware} and then introduce the software architecture in Sec.~\ref{sec: modules}. 

\subsection{Robot}   \label{sec: hardware}

We build our robot system based on the Rover mobile robot~\cite{rover_robotics_rover_2023}. As shown in Fig.~\ref{fig: hardware}, we install two Realsense cameras to perform navigation in the indoor environment and simultaneously track human pose. 
More specifically, we use a Realsense D455 as our front camera and use a Realsense D435 as the rear camera. We map and localize the robot with the front camera by combining visual odometry and IMU sensor reading. 
Our onboard computer is equipped with an Intel I7 processor, 32G RAM, and a P3200 Nvidia GPU.

\subsection{Modules}    \label{sec: modules}

Our mobile robot system consists of four modules built on the Robot Operation System (ROS)~\cite{quigley_ros_2009}. We briefly describe our system architecture in this section; additional details can be found in the supplementary material.


\textbf{Mapping and localization.} To create 2D and 3D maps, we use the RGB-D image and IMU data from the front camera. We apply the \textit{rtabmap} package~\cite{labbe_rtab-map_2019} to create visual odometry from color images and merge it with the IMU reading by using the \textit{robot\textunderscore localization} package~\cite{moore_generalized_2014}. 
The merged odometry is used for mapping the environment and for localization during navigation.

\textbf{Human Pose Detection.} We use \textit{Yolo-v8}~\cite{jocher_yolo_2023} to extract the human skeleton pose from the color images and set the detection frequency to 15Hz on our onboard computer. By combining the pose estimation result with the depth image and the localization result, we obtain the 3D skeleton pose and transform it into the global frame.

\textbf{Human Pose Forecast.} We run our human pose forecasting algorithm in real-time as in Sec.~\ref{sec: approach}. We collect consecutive 3D poses during inference and transform them into the latest pose frame. In practice, we execute the forecast in $15$-fps and set our predicting horizon as $3$ seconds.

\textbf{Navigation.}
We formulate the path planning task as a finite-horizon optimal control problem.
We define our objective function at each time step based on the Pixels-Per-Area (PPA) metric~\cite{jiang_onboard_2023}. PPA measures the viewing quality by considering the viewing distance and viewing angle. 
Given the predicted human motion, we calculate a sequence of robot controller inputs such that the total PPA cost over-time is maximized.
We solve this optimal control problem using the Dynamic Programming (DP)~\cite{bellman_dynamic_1966} method and plan the robot's trajectory.
A sequence of target robot poses is calculated and fed as input to the \textit{move\textunderscore base} package~\cite{marder-eppstein_office_2010}. 
A local path is planned to avoid collision with the environment and is executed by sending velocity commands to the robot. 
We provide more details on this formulation in the supplementary material attached to our project website.

%% file: sections/06-dataset.tex

In addition to the standard synthetic dataset: GTA-IM~\cite{cao_long-term_2020}, we are also interested in evaluating our method on a large-scale real-world dataset. However, existing realistic datasets, such as PROX~\cite{hassan_resolving_2019}, are limited to single-room areas. Therefore, we collect and present the Real Indoor Motion (Real-IM) dataset for the large-scale human follow-ahead. We simultaneously capture the entire human body and environment in building-scale spaces using the robot described in Sec.~\ref{sec: hardware}. Compared to the existing synthetic dataset (GTA-IM), our dataset contains more movement patterns, including walking, crab moving, and varying moving speeds.

We collect 12 sequences from 5 different building halls. Each sequence contains an approximately four-minute movement. The dataset includes sequences with different lighting conditions recorded at different times of day. We invited multiple actors with different walking styles and genders. 
We provide the raw ROS bags and pre-processed data for direct use. 
In the supplementary material from our webpage, we present a few sample images from our dataset visualized in Rviz~\cite{quigley_ros_2009}.

%% file: sections/07-experiments.tex
\begin{figure*}[h!]
    \centering
        \includegraphics[trim={6cm 5.2cm 6cm 1.7cm},clip,width=\textwidth]{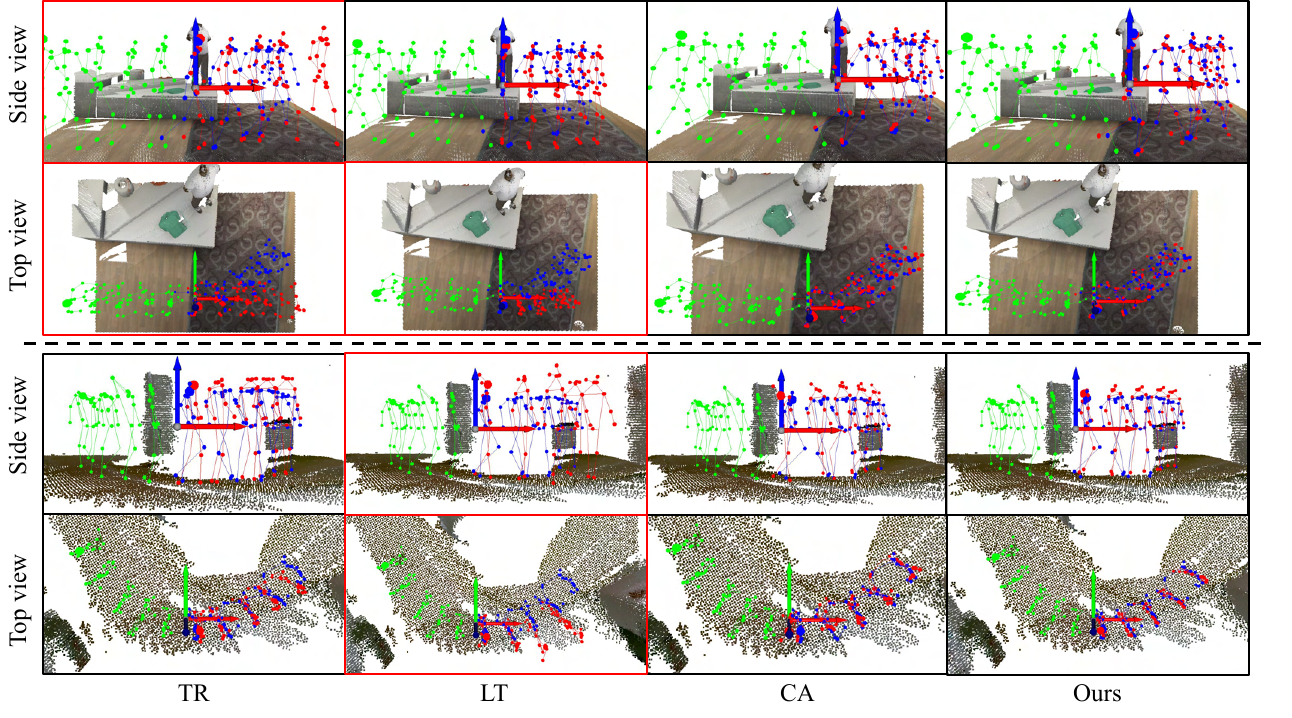}
    \caption{\textbf{Qualitative results of the human pose forecasting.} Each column corresponds to a different method. Human pose history, prediction results, and ground truth are shown in green, red, and blue. 
    TR~\cite{vaswani_attention_2017} and LT~\cite{cao_long-term_2020} fail to predict the turn  (red outline), whereas CA~\cite{mao_contact-aware_2022} and ours are successful in correctly predicting it (black outline).}
    \label{fig: qualitative}
    \vspace{-5pt}
\end{figure*}

In this section, we present results from experiments to compare our human pose prediction algorithm with existing works. We introduce the experiment setup (Sec.~\ref{sec: setup}), results (Sec.~\ref{sec: exp_pose}), and analysis (Sec.~\ref{sec: ablation}).
We will then demonstrate the robot follow-ahead task with our algorithm and justify the need for our human pose prediction for this task (Sec.~\ref{sec: exp_robot}). 

\subsection{Experiment setup}  \label{sec: setup}

\textbf{Baselines.}   \label{sec: baselines}
We include two human pose forecasting baselines: CA~\cite{mao_contact-aware_2022} and LT~\cite{cao_long-term_2020}. Similar to our setting, these methods utilize environmental information when predicting 3D human pose. 
LT~\cite{cao_long-term_2020} is the first work to forecast human poses by sequentially predicting a trajectory and estimating the joint poses.
Similar to our network architecture, CA~\cite{mao_contact-aware_2022} applies GRU for human pose prediction. 
As mentioned in the CA~\cite{mao_contact-aware_2022} paper, CA outperforms LTD~\cite{mao_learning_2019}, DMGNN~\cite{li_dynamic_2020}, and SLT~\cite{wang_synthesizing_2021}. Therefore, we do not include these methods in our experiments. In addition to CA and LT, we implement a pure Transformer-based method (indicated as TR) by taking the history of human poses as input and output predicted future poses. 

\textbf{Metrics.}
We use Mean Per Joint Position Error (\textbf{MPJPE}) to measure our performance in predicting the pose and the trajectory, which is a standard and widely used metric in human pose prediction~\cite{lyu_3d_2022}. 
MPJPE provides a quantitative measure of how close the predicted joint positions are to the true joint positions, averaged across all joints. A lower MPJPE indicates a better model performance.
As in~\cite{mao_contact-aware_2022} and~\cite{cao_long-term_2020}, we report hip position errors as the global transition error (path error) and local 3D pose errors in millimeters (mm).

\textbf{Implementation details.}
We sequentially train our PathNet and PoseNet for 1200 epochs each using Adam~\cite{kingma_adam_2017} optimizer implemented in Pytorch~\cite{paszke_pytorch_2019}.  The learning rate is set to $0.001$ for the PathNet and $1e-5$ for the PoseNet. The learning rate scheduler has a gamma of $0.1$ and a step size of $600$. On the GTA-IM dataset, we set the trajectory map and occupancy map size as $40 \times 40$ and use them to represent $5m^2$ local space. We choose the same training and testing set as~\cite{mao_contact-aware_2022}. On the Real-IM dataset, we use a forecasting horizon of $3$s and predicted occupancy map and trajectory map of size $8m^2$.

\subsection{Human Pose Prediction}  \label{sec: exp_pose}

\begin{table*}[htbp]
    \small
    \caption{\textbf{Evaluation results in GTA-IM dataset}. We use the same training and testing set as CA~\cite{mao_contact-aware_2022} and use their reported numbers. Results show that our methods outperform the state-of-the-art methods. The results also show that by providing additional map information, trajectories are predicted significantly better, which yields lower pose errors.}
    \begin{center}
        \begin{tabular} {rccccccccc} 
            \toprule
            &\multicolumn{4}{c}{\bf Path error (mm)}    &\multicolumn{4}{c}{\bf 3D Pose error (mm)}     \\
            \cmidrule{2-5} \cmidrule{6-10}
            \bf Time (sec)                  &0.5            &1.0            &1.5            &2.0        &0.5            &1.0            &1.5            &2.0            &mean (2s)   \\
            \midrule
            TR~\cite{vaswani_attention_2017} &113.7          &187.4          &375.8          &471.2      &112.4          &116.4          &129.3          &139.8          &118.3          \\
            LT~\cite{cao_long-term_2020}     &104            &163            &219            &297        &91             &158            &237            &328            &173            \\
            CA~\cite{mao_contact-aware_2022} &58.0           &103.2          &154.9          &221.7      &\textbf{50.8}  &67.5           &75.5           &86.9           &\textbf{61.4}  \\
            Ours (PathNet+GRU)              &\textbf{52.5}  &\textbf{99.6}  &\textbf{107.3} &\textbf{113.8} &51.1       &\textbf{63.6}  &\textbf{70.7}  &\textbf{75.0}  &62.9           \\ 
            \midrule
            PathNet (partial)                & 123.8 &149.3  &237.0  &290.9 & 72.9  &83.5  &114.7  &126.9  &110.4\\
            PathNet (unknown)                & 120.5 & 145.1 & 232.6 &286.1  & 72.4 &82.8  &105.5  &118.0  &83.5\\
            \midrule
            Ours w/o $\mathcal{L}_{col}$    & 112.8 &153.6  &243.4  &292.0 & 71.6 &81.0   &100.6  &110.6  &81.0\\
            Ours w/o $\mathcal{L}_{map}$    & 132.0 &184.4  &320.1  &365.1  & 73.9 &86.3  &129.6  &146.1  &93.3\\
            \bottomrule
        \end{tabular}
    \end{center}
    
    \label{tab: GTA-IM_results}
\end{table*}

\begin{table*}[!htbp]
    \centering
    \small
    \caption{\textbf{Evaluation results in Real-IM dataset}. We report results across both 2s and 3s prediction horizons. Our method performs better than the baselines on both path and pose prediction. As expected, the error increases along with the horizon for all methods. }
    \begin{tabular} {@{\extracolsep{-2pt}}rcccccccccccc@{}} 
        \toprule
        &   \multicolumn{6}{c}{\bf Path error (mm)}                             &\multicolumn{6}{c}{\bf 3D Pose error (mm)} \\
        \cmidrule{2-7} \cmidrule{8-13}
        \bf Time (sec)                  &1.0    &1.5    &2.0    &3.0    &mean (2s)    &mean (3s)    &1.0    &1.5    &2.0    &3.0    &mean (2s)    &mean (3s) \\
        \midrule
        TR~\cite{vaswani_attention_2017}&150.5  &249.1  &295.7  &439.1  &154.3      &240.3      &89.9   &91.8   &93.8   &101.2  &89.8   &92.4   \\
        CA~\cite{mao_contact-aware_2022}&153.9  &255.7  &257.6  &374.3  &253.2      &257.4      &\textbf{67.9}   &80.3   &86.4   &109.2  &69.7   &80.7   \\
        \midrule
        Ours (PathNet+TR) & \multirow{2}{*}{\textbf{145.8}} & \multirow{2}{*}{\textbf{165.7}} & \multirow{2}{*}{\textbf{186.9}}  & \multirow{2}{*}{\textbf{240.2}}  & \multirow{2}{*}{\textbf{150.9}}      & \multirow{2}{*}{\textbf{178.4}}     &96.8   &97.0   &98.1   &98.4   &96.5   &97.1   \\      
        Ours (PathNet+GRU)&       &       &       &       &           &           &69.9   &\textbf{75.5}   &\textbf{78.1}   &\textbf{84.5}   &70.7   &\textbf{75.0}   \\
        \midrule
        PathNet (partial)                &202.0  &319.5  &372.4  &511.5  &251.9  &350.7  &80.9   &88.6   &92.7   &100.8  &83.2   &92.9  \\
        PathNet (unknown)                &183.3  &316.7  &390.9  &583.2  &192.7  &308.8  &69.3   &76.6   &81.2   &101.9  &\textbf{69.3}   &76.5  \\
        \midrule
        Ours w/o $\mathcal{L}_{col}$    &204.4  &449.6  &341.0  &493.2  &418.0  &440.6  &101.5  &102.3  &95.2   &115.0  &99.8   &107.0  \\
        Ours w/o $\mathcal{L}_{map}$    &162.1  &271.8  &441.4  &763.3  &171.9  &316.1  &68.5   &80.4   &92.9   &132.0  &69.8   &90.5  \\
        \bottomrule
    \end{tabular}
    
    \label{tab: Real-IM_results}
\end{table*}

We compare our human pose prediction network with the baselines on both synthetic and the proposed real-world datasets. As shown in Table~\ref{tab: GTA-IM_results} and Table~\ref{tab: Real-IM_results}, our method outperforms the baselines on both trajectory prediction and human pose forecasting. 
We improve path prediction significantly by providing additional map information beyond visibility in first-person view.
On the pose prediction side, our method achieves lower error after the first second, which indicates that CA~\cite{mao_contact-aware_2022} is more accurate for short-term prediction, while ours is better for longer forecasting horizons, taking advantage of the trajectory prediction accuracy.


\textbf{Inference time.} In addition to accuracy, we compare inference speed on our onboard computer across all methods. Since CA~\cite{mao_contact-aware_2022} considers pair-wise point features between every 3D point and human joint across multiple time frames, it takes an average of $100.13ms$ to predict the human poses. In contrast, considering only 2D maps, our method has an average inference time of $32.12ms$, allowing us to respond faster in real-world applications.

\subsection{Ablation Study}  \label{sec: ablation}
In this section, we investigate the impact of different components in our network on human pose prediction. 

\textbf{Map visibility.} One of our key contributions is to provide the local environment map to our PathNet. Therefore, we study how different map visibility can affect the pose prediction accuracy. We compare our results, which utilize a fully known local map, with two baselines: 1) without access to the map and 2) with a map whose visible area is limited to the camera FOV, as if the robot is operating in a new environment. Results from Table~\ref{tab: GTA-IM_results} and ~\ref{tab: Real-IM_results} show that map information does play a critical role in improving prediction accuracy. 

\textbf{GRU vs. Transformers.} Transformers~\cite{vaswani_attention_2017} have been shown to outperform GRUs when there's a sufficient amount of data~\cite{lucas_posegpt_2022}. In this section, we compare the performance of these two modules by replacing GRU with Transformers in the PoseNet module. As shown in Table~\ref{tab: Real-IM_results}, we observe that GRU is outperforming Transformers on the 3D pose prediction task. One possible reason is that our training set is not large enough, and GRU is sufficient in this case.

\textbf{Loss terms.} 
During the training process, we expect that by using $\mathcal{L}_{col}$ and $\mathcal{L}_{map}$, the predicted trajectory would avoid collision with the environment. As shown in Table~\ref{tab: GTA-IM_results} and ~\ref{tab: Real-IM_results}, we conduct an ablation study to investigate whether these loss terms are helpful. Results show that these collision loss terms do increase the accuracy of both short-term and long-term trajectory prediction.

\subsection{Robot Follow-ahead}  \label{sec: exp_robot}
\begin{figure}[th]
    \begin{center}
        \includegraphics[width=1.0\columnwidth]{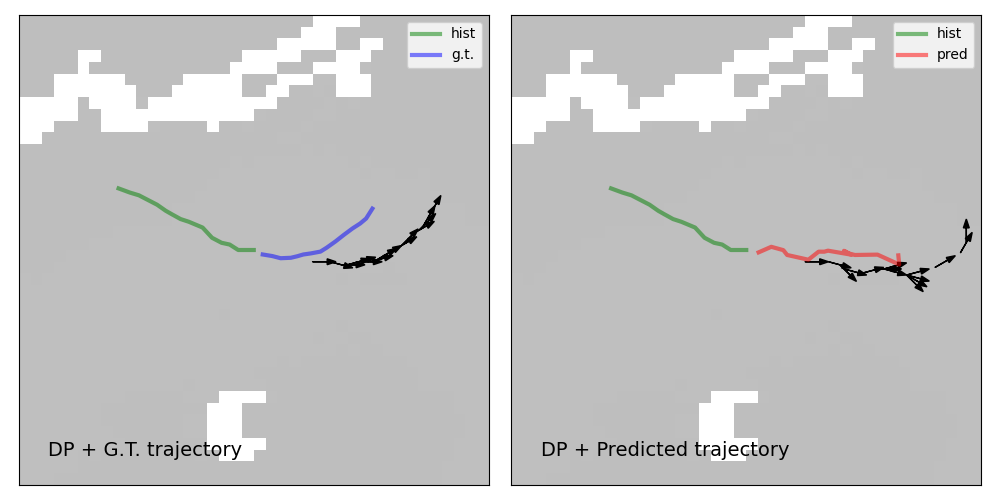}
    \end{center}
    \caption{\textbf{Robot Follow-ahead}. We visualize the planned path based on the human pose predictions. The human trajectory history, trajectory prediction, and the ground truth are shown in green, red, and blue. We visualize the planned robot path in arrows for each time step. We visualize the map as the background, white for obstacles and gray for free space.}
    \label{fig: follow-ahead}
\end{figure}

In this section, we investigate the performance of our algorithm combined with the robot system on the robot follow-ahead task. As shown in Fig.~\ref{fig: follow-ahead}, we compare the planned robot path using the predicted trajectory against using the ground truth. We present additional qualitative results in the supplementary video.

Moreover, we investigate whether our human pose prediction helps with the robot follow-ahead task.
We compare our controller, with the human pose prediction as the input, against two other controlling strategies: 
(1) a greedy approach implemented with Extended Kalman Filter (EKF) for estimating human position: 
We do not apply any future pose prediction. Instead, we use a uniform Gaussian distribution to propagate a future trajectory from the past human trajectory.
We demonstrate the performance of a myopic controller that directly reacts to the EKF output. 
(2) a Dynamic Programming-based optimal controller that is given a ground-truth future trajectory as part of the input (DP + g.t. traj.): 
This oracle algorithm serves as an upper-bound controller of this task.


We evaluate the robot's follow-ahead performance using the following metrics. 1) \textbf{Area}. Percentage of pixels in images that the actor occupies. 2) \textbf{Tracking time}. The number of frames in which the actor is detected in the image is divided by the total number of frames in a sequence. 3) \textbf{Distance}. The distance between the center of the human bounding box and the center of the image is normalized by the image width.

Experiment results are shown in Table~\ref{tab: Follow-ahead_results}. 
Even though our algorithm does not have access to the future trajectory of the human, with its predictive capabilities, its performance is better than the myopic controller and is within 85\% of this oracle-based upper-bound (Table~\ref{tab: Follow-ahead_results}).
The result shows that the robot follow-ahead performance can benefit from the human pose forecasting predictions. 

\begin{table}[th]
    \caption{\textbf{Follow-Ahead Evaluation.}}
    \begin{center}
        \begin{tabular} {@{\extracolsep{2pt}}lccc@{}} 
            \toprule
            \bf Method  &Area $\uparrow$   &Tracking Time$\uparrow$    &Distance$\downarrow$  \\
            \midrule
            LB: EKF         &0.241              &0.67           &0.147          \\
            Ours: DP + pred. traj.        &0.302              &0.85           &0.151          \\
            UB: DP + g.t. traj.         &\textbf{0.358}     &\textbf{1.00}  &\textbf{0.145} \\
            \bottomrule
        \end{tabular}
    \end{center}
    
    \label{tab: Follow-ahead_results}
\end{table}

%% file: sections/08-conclusions.tex
This paper presented a method to predict human poses in indoor environments in order to accomplish the robot follow-ahead task. 
We proposed an architecture that first predicts the 2D human trajectory based on the occupancy map and then predicts the 3D human poses conditioned on the 2D trajectory.
To validate our approach, we built a mobile robot system and collected a building-scale Real Indoor Motion Dataset (Real-IM) for human pose forecasting problems in large and complex environments. 
Through both synthetic and realistic experiments, we showed that our approach outperforms baselines and state-of-the-art methods on trajectory prediction and long-term human pose forecasting. In terms of run-time, it is three times faster. 
We also demonstrated successful robot follow-ahead by forecasting human poses in real-time. 

Our system has its limitations. In general, most failure cases of the robot follow-ahead task stem from the rear camera losing track of the actor. Consequently, the robot is unable to locate the actor and to move to a position such that the actor remains within sight. 
On the human pose prediction task, we observe that our human pose prediction method does not guarantee consistency across time and may lead to jerky motion under challenging scenarios, such as T-junctions. We provide some examples of these failures in the supplementary video.
Meanwhile, we highlight that the planning algorithm for this robot follow-ahead problem can be further developed to incorporate the rear camera's limited field-of-view (FOV) and constrained robot kinematics.
We plan to address these challenges in our future work. 


%% file: sections/supplementary/intro.tex
Our main paper proposes a new human pose prediction method for the robot follow-ahead (RFA) problem. Through experiments, we show that our proposed method performs better or is comparable to the state-of-the-art methods while computing three times faster. Meanwhile, we implement our algorithm on a real robot, showing that the robot system can benefit from our human pose prediction results and complete the robot follow-ahead task with a certain successful rate.

Although the controller side of the robot is not our primary focus, we find it non-trivial to the success of the tasks. In the supplementary material, we would like to provide additional insight into the problem of robot follow-ahead (Sec.~\ref{appendix: follow-ahead}). 
Meanwhile, we will provide more technical details (Sec.~\ref{appendix: robot}) on the mobile robot. 
We provide samples on the Real-IM dataset in Sec.~\ref{appendix: dataset}.

%% file: sections/supplementary/follow-ahead.tex
In this section, we formulate the robot follow-ahead (RFA) task as an optimal control problem and provide some insight into its difficulty, which we plan to address in future work.

\subsection{Problem Formulation}
Following our formulation in our main paper (Sec.~III), given the predicted human poses $\mathbf{\hat{X}}_{N+1:N+T}$, we formulate the robot follow-ahead control problem as below. For simplicity, we use $(\cdot)_{:N}$ and $(\cdot)_{N:}$ to denote the past and the future.

We define the robot state $\mathbf{Y}_t$ at time $t$ by its 2D position $x$, $y$ and its yaw angle $\theta$. $\mathbf{Y} = (x, y, \theta)$. We define the robot control command with linear and angular velocity $\mathbf{u} = (v, \omega)$.
We write the transition function as Eq.~\ref{appendix: eq: transition}.
\begin{equation}    \label{appendix: eq: transition}
    \begin{split}
        &\mathbf{Y}_{t+1} = \mathbf{Y}_t + \mathbf{B} \mathbf{u} \delta t   \\
        &\mathbf{B} = 
            \begin{bmatrix}
                \cos(\theta)    & 0\\
                \sin(\theta)    & 0\\
                0               & 1\\
            \end{bmatrix}
        \quad
        \mathbf{u} = 
            \begin{bmatrix}
                v   \\
                \omega
            \end{bmatrix}
    \end{split}
\end{equation}

We define the cost function $\mathcal{J}$ as a function of the state and the map, measuring the viewing quality and all the other considered costs. 
In our case, for example, we define the view quality cost related to Pixels-Per-Area (PPA)~\cite{jiang_onboard_2023}, described as $\mathcal{L}_{view} = \frac{\|\mathbf{Y}_t - \mathbf{X}_t\|}{\cos(\gamma)}$. $\gamma$ is defined by the angle between the camera focal direction and the human heading direction. We also define the cost of collision with the environment.

\begin{equation}
    \mathcal{J} = \sum_{t} \left(\mathcal{L}_{view} + \lambda_{col} \mathcal{L}_{col} \right)
\end{equation}

Given the cost function $J$, we calculate a sequence of robot control inputs $\{\mathbf{u}_{N+1}, \ldots, \mathbf{u}_{N+T}\}$, such that the cost is minimized.

\begin{equation}
    \mathbf{u}_{N:} = \arg\min_{\mathbf{u}} \mathcal{J}
\end{equation}


Our main paper uses a dynamic programming (DP) method to solve this optimal control problem. We build the value function table with dimension $H \times W \times \Theta \times T$, where $H$, $W$, $\Theta$ is the discretized dimension of the robot state on $x$, $y$ and $\theta$ respectively. However, this DP method is computationally costly and only works when discretizing the state and action spaces in a coarse division. 
We plan to implement other methods, such as Model Predicted Control (MPC) \cite{camacho_constrained_2007}, to solve this problem in future work.

\subsection{Challenge}
We highlight some of the difficulties we encountered in the real-robot setting when solving this robot's follow-ahead problem.

\subsubsection{\textbf{Robot follow-ahead (RFA) v.s. tracking}}
\begin{figure}[ht]
    \begin{center}
        \includegraphics[width=0.9\columnwidth]{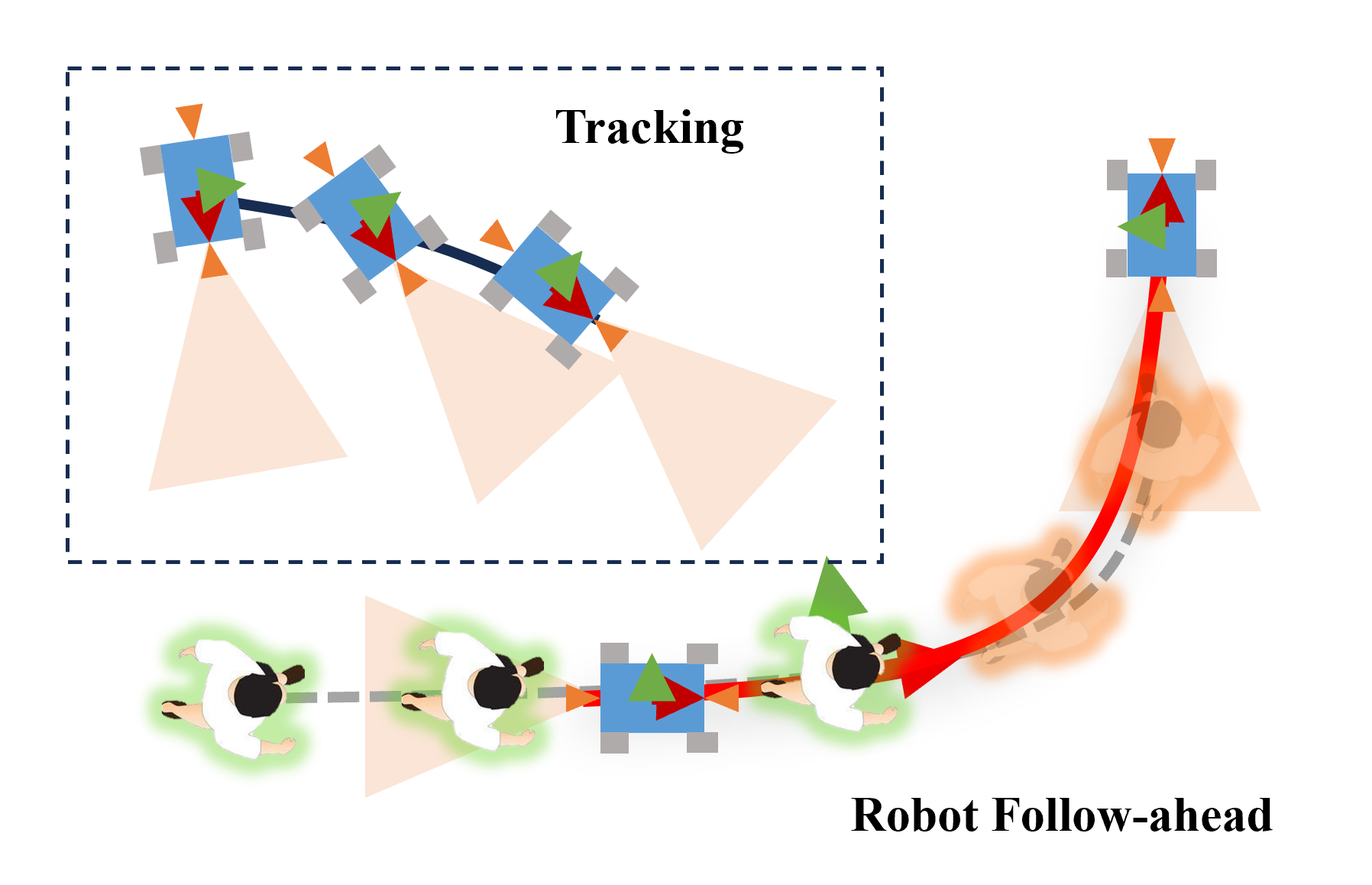}
    \end{center}
    \caption{The robot follow-ahead problem, unlike the classical passive tracking problem, uses an extra rear camera and acts ahead of the human. Meanwhile, handling human rotation while being in front is more challenging.}
    \label{fig: supplementary: rfa_vs_tracking}
\end{figure}

As a fundamental comparison, RFA is more challenging than classical tracking problem for the following reasons (Fig.~\ref{fig: supplementary: rfa_vs_tracking}.
1) RFA uses an extra rear camera to track the human. 
2) RFA makes decisions earlier than the human: in an indoor hallway environment, the robot reaches the turn before the human does and, therefore, has to make a prediction and respond to the turn earlier than the human.
3) RFA moves a longer distance to the human in-place rotation: the robot needs to move circularly through a long distance, whereas the robot stays still in the classical tracking problem.

\subsubsection{\textbf{Human omnidirectional motion v.s. constrained robot motion and limited Field-of-View (FOV)}}
\begin{figure}[ht]
    \begin{center}
        \includegraphics[width=0.9\columnwidth]{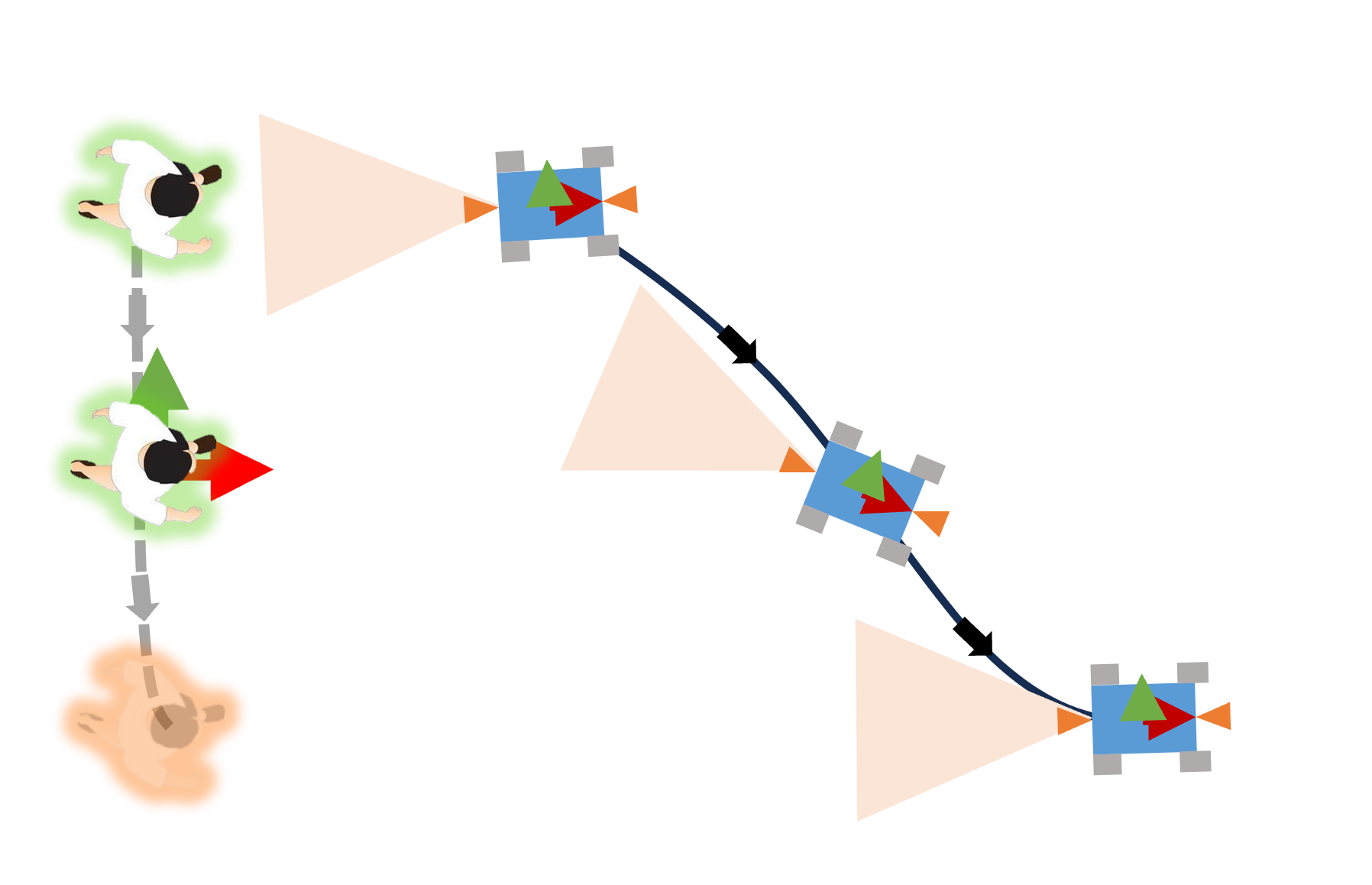}
    \end{center}
    \caption{The human heading direction could differ from the motion direction, bringing an extra challenge to the robot follow-ahead problem with the limited FOV.}
    \label{fig: supplementary: rfa_hde}
\end{figure}

Another challenge of this robot follow-ahead problem is that a human can move omnidirectionally while the robot (and the rear camera) is constrained by its kinetics. Moreover, the limited camera's field-of-view (FOV) makes it more challenging to keep track of the human. We show an example in Fig.~\ref{fig: supplementary: rfa_hde}, where the human moves in the side direction. 
While the robot attempts to execute the path and maintain its position in front, the rear camera films in the opposite direction (due to the coupling constraints by the robot motion), suffers from a limited FOV, and can easily lose track of the human.

\subsubsection{\textbf{Time consistency}}
As we discussed in Sec. VIII, the current method of human pose prediction does not guarantee consistency over time. Therefore, if the prediction results vary sharply in a short period, the controller may overreact, causing the robot to move jerkily or get stuck at the turn. This happens more often in a T-junction or cross-junction, where the robot has to select among directions without enough prior information. We show demonstration of these failed cases in the supplementary video.




%% file: sections/supplementary/robot.tex
\begin{figure*}[htbp]
    \centering
    \includegraphics[width=0.9\textwidth]{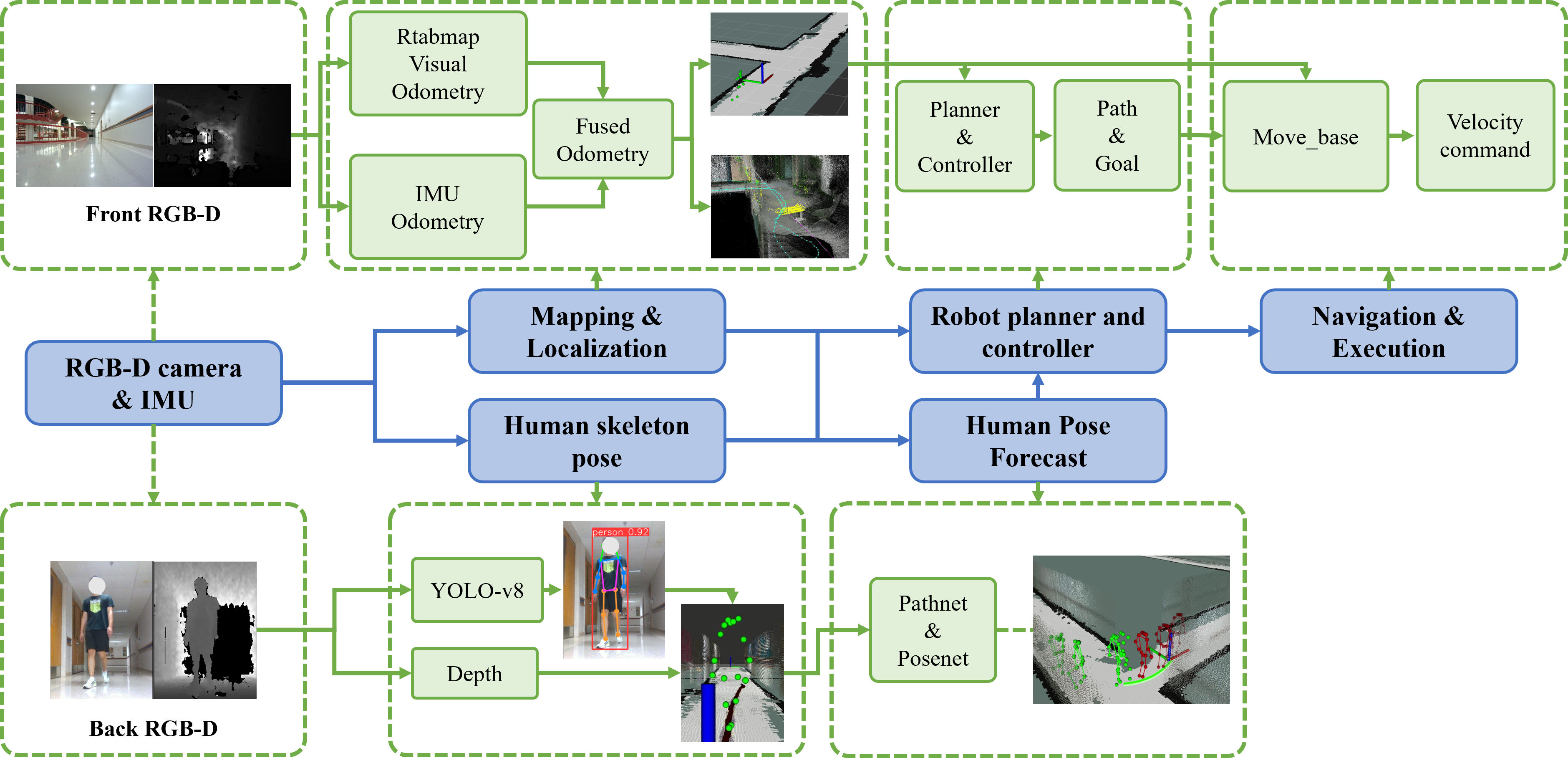}
    \caption{\textbf{Modules}. Our system architecture's main components are in blue, and sub-modules in green. RGB-D images are used to localize the robot and extract human skeleton poses. Human poses are predicted. The robot path is planned and executed afterward.}
    \label{fig: modules}
\end{figure*}

In this section, we provide detailed information on our robot system for the robot follow-ahead task. 
In Fig.~\ref{fig: modules}, we show the architecture of our robotic system. We explain the main modules and their functionality in our main paper. In this section, we want to provide more technical details to help replicate our work.


\subsection{Cameras setting}
We use our front camera (Realsense D455) to localize the robot and a rear camera (Realsense D435) to track the 3D human skeleton poses. 
We install the rear camera vertically with a $5^{\circ}$-tilting angle to capture the entire human body and track the human actor. 
We set the front camera resolution at $15$Hz with $640\times480$ resolution. We set the rear camera at $30$Hz with $1280\times720$ resolution. We clip distance by $6$m to remove noise from the background. We merge visual odometry from \textit{rtabmap} \cite{labbe_rtab-map_2019} and IMU odometry from Realsense D455 with \textit{robot\textunderscore localization} \cite{moore_generalized_2014} package.

\subsection{Mapping and Localization}
Although Simultaneous Localization and Mapping (SLAM) in an indoor environment has been studied for years, we find it difficult to find any open-sourced examples in our setting.
We provide detailed parameters for the mapping and localization using \textit{rtabmap} in Table~\ref{tab: rtabmap}, which we spent much effort and view as of much value. 
These parameters perform well in indoor environments in general and are tuned to fit some featureless environments such as hallways.
All the environmental information is recorded as the point cloud. The reconstruction results perform well on the loop closure.

\begin{table}[thp]
    \centering
    \small
    \caption{\textbf{Mapping and Localization.} Parameters for the robot mapping and localization using \textit{rtabmap} and \textit{robot\textunderscore localization}. From our hands-on experience, these works best in the indoor environment.}
    \begin{tabular}{llc}
        \toprule
        Package                     &Parameters                             &Value  \\
        \midrule
        \multirow{13}{*}{rtabmap}   &Grid/NoiseFilteringMinNeighbors        &5      \\
                                    &Grid/NoiseFilteringRadius              &0.1    \\
                                    &Grid/MaxGroundHeight                   &0.2    \\
                                    &Grid/MaxObstacleHeight                 &2.0    \\
                                    &Grid/RangeMax                          &6.0    \\
                                    &Grid/RangeMin                          &0.6    \\
                                    &Reg/Force3DoF                          &true   \\
                                    &Reg/Strategy                           &0      \\
                                    &Kp/MaxDepth                            &5.0    \\
                                    &Mem/RehearsialSimilarity               &0.30   \\
                                    &Rtabmap/DetectionRate                  &15     \\
                                    &Vis/MinInliers                         &8      \\
                                    &Vis/MeanInlierDistance                 &1.0    \\
        \midrule
        \multirow{3}{*}{rgbd\textunderscore odometry} &Reg/Force3DoF        &true   \\
                                    &Reg/Strategy                           &0      \\
                                    &Vis/MinInliers                         &8      \\
        \midrule
        \multirow{6}{*}{ukf\textunderscore se}  &frequency                  &30     \\
                                    &two\textunderscore d\textunderscore mode   &true   \\
                                    &odom\textunderscore relative           &true  \\
                                    &imu0\textunderscore differential       &true   \\
                                    &imu0\textunderscore relative           &false  \\
                                    &use\textunderscore control             &false  \\
        \bottomrule
    \end{tabular}
    \label{tab: rtabmap}
\end{table}

\begin{figure}[thbp]
    \centering
    \includegraphics[width=1.0\columnwidth]{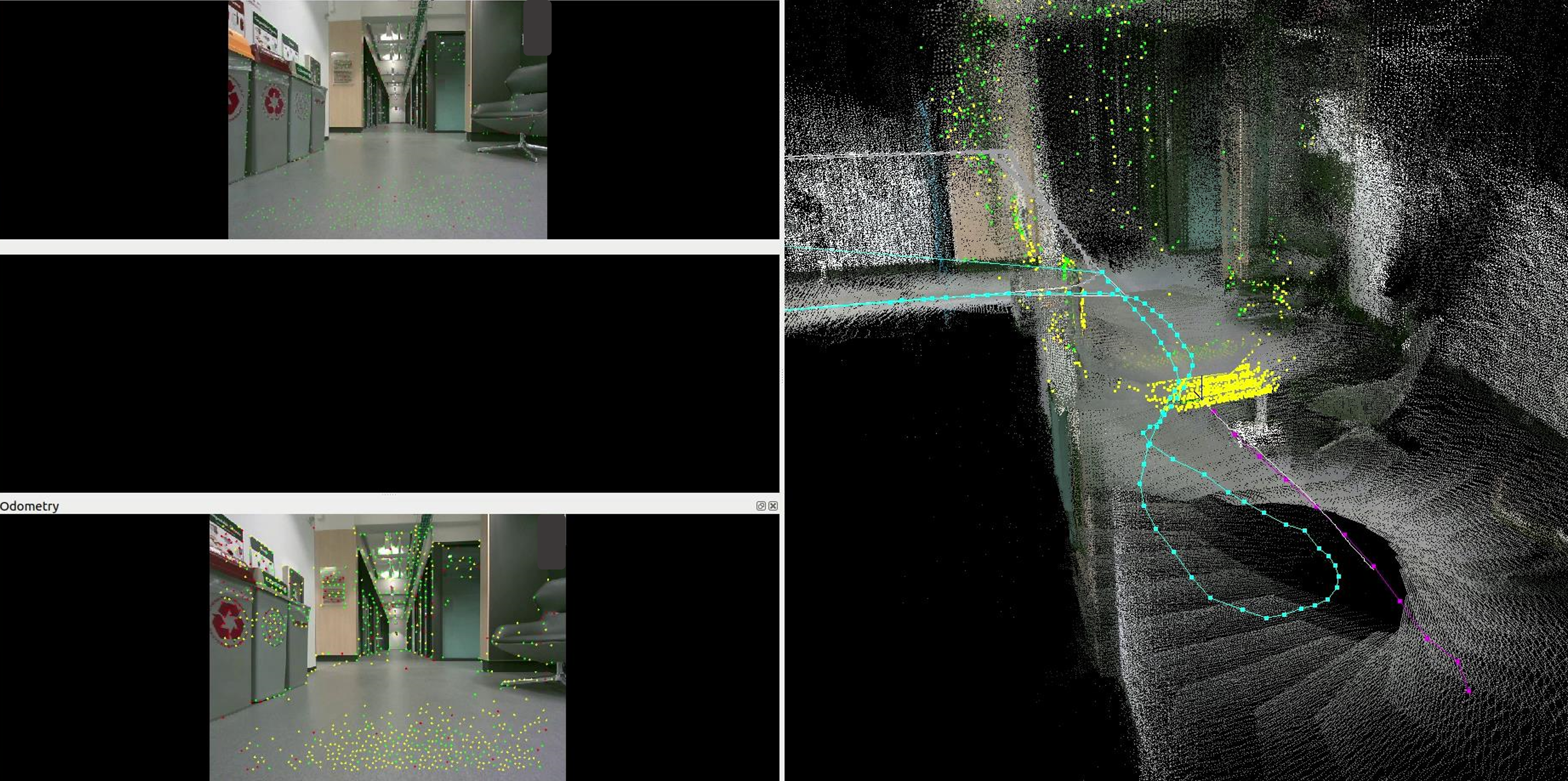}
    \caption{\textbf{\textit{Rtabmap} visualization}. Left top: the raw image input from the front camera. Left bottom: extracted visual features. Right: The built 3D map as the point cloud. Blue line: the path our robot executed when building the map. Each node in the blue line represents an image sample stored in the database when localizing the robot.}
    \label{fig: rtabmapvis}
\end{figure}

In Fig.~\ref{fig: rtabmapvis}, we show the extracted image feature and localized odometry when building the map in the \textit{Rtabmapviz} visualization tool. During the localization process, visual features are extracted and compared against the stored image features from the database.

\subsection{Navigation}
We also provide part of the \textit{move\textunderscore base} parameters in the \textit{navigation\textunderscore stack} \cite{marder-eppstein_office_2010} in Table~\ref{tab: movebase}. Parameters in Table~\ref{tab: movebase} need to be tuned for different robot setups.

\begin{table}[ht]
    \centering
    \small
    \caption{\textbf{Navigation.} Parameters for the robot navigation with the built map using \textit{move\textunderscore base} package.}
    \begin{tabular}{llc}
        \toprule
        Package                     &Parameters                             &Value  \\
        \midrule
        \multirow{4}{*}{global\textunderscore costmap} 
                                    &update\textunderscore frequency        &10     \\
                                    &publish\textunderscore frequency       &10     \\
                                    &footprint\textunderscore padding       &0.05   \\
                                    &inflation\textunderscore layer/inflate\textunderscore unknown        &false  \\
        \midrule
        \multirow{5}{*}{costmap\textunderscore common} 
                                    &obstacle\textunderscore range          &2.5    \\
                                    &raytrace\textunderscore range          &0      \\
                                    &robot\textunderscore radius            &8      \\
                                    &inflation\textunderscore radius        &0.1    \\
                                    &transform\textunderscore tolerance     &1.0    \\
        \bottomrule
    \end{tabular}
    \label{tab: movebase}
\end{table}

%% file: sections/supplementary/dataset.tex
We provide a few samples of our Real Indoor Motion (Real-IM) dataset. Each sequence contains the RGB-D readings from the front camera (used for the SLAM) the rear camera (used for extracting human poses), and the corresponding built map database. Usage of the database and the code can be found on our webpage. \url{https://qingyuan-jiang.github.io/iros2024\_poseForecasting/}.

\begin{figure}[!th]
    \begin{center}
        \includegraphics[width=1.0\linewidth]{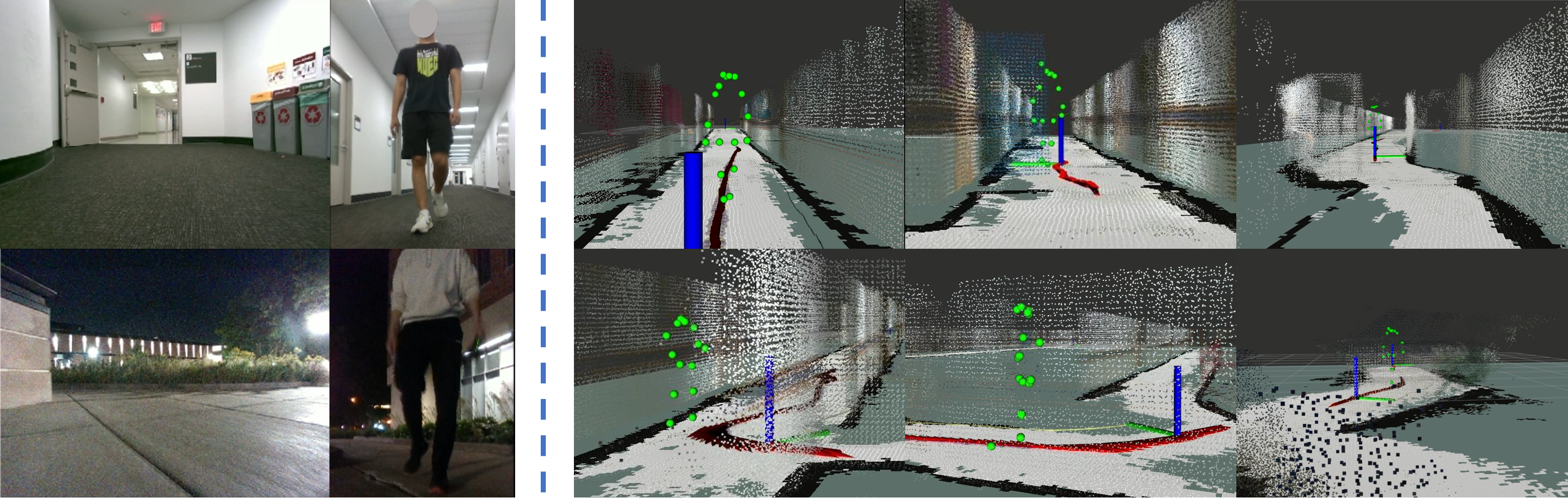}
    \end{center}
    \caption{\textbf{Real Indoor Motion (Real-IM) Dataset samples}. We collect a building-scale human motion dataset using the mobile robot. We record full 3D environment point clouds and build the corresponding occupancy map. The two columns on the left are the raw front and rear camera observations. The right three columns show the actor's pose in the reconstructed environment. }
    \label{fig: dataset}
\end{figure}